\pdfoutput=1

\documentclass[11pt]{article}

\usepackage{acl}
\usepackage{times}
\usepackage{latexsym}
\usepackage{enumitem}
\usepackage{hyperref}
\usepackage[T1]{fontenc}

\usepackage[utf8]{inputenc}

\usepackage{microtype}

\usepackage[ruled,vlined]{algorithm2e}  
\usepackage{float}
\usepackage{algpseudocode}
\usepackage{multirow}
\usepackage{tabularx}
\usepackage{times}
\usepackage{threeparttable}
\usepackage{latexsym}
\usepackage{tikzsymbols}
\usepackage{amsthm}
\usepackage{adjustbox}
\usepackage[section]{placeins}
\usepackage{graphicx}
\graphicspath{ {./diagrams/} }
\usepackage{amsmath}
\usepackage{booktabs}
\usepackage{microtype}
\usepackage[T1]{fontenc}

\usepackage[utf8]{inputenc}
\usepackage{xspace}
\newcommand{\dataset}{\texttt{ViLPAct}\xspace}
\newcommand{\charades}{\texttt{Charades}\xspace}

\newcommand{\bert}{\textsc{BERT}\xspace}

\newcommand{\uniVL}{\textsc{Act-UniVL}\xspace}
\newcommand{\prophetNet}{\textsc{Act-ProphetNet}\xspace}
\newcommand{\prophetNetEN}{\textsc{ProphetNet-EN}\xspace}
\newcommand{\itd}{\textsc{I3D}\xspace}
\newcommand{\bmTF}{\textsc{BM25}\xspace}

\newcommand{\act}[1]{\textit{#1}\xspace}
\newcommand{\probInf}{\textit{ProbInf}\xspace}

\newcommand\RetrievalScoring{NSPlan\xspace}

\newcommand\UniVLProphetNet{TwoStagePlan\xspace}

\usepackage{amsthm,amsmath,amsfonts,bm,xspace}
\usepackage{color}

\def\vs{{\em v.s.}\xspace}

\def\eqref#1{(\ref{#1})}

\def\1{\bm{1}}

\def\va{{\bm{a}}}

\def\vg{{\bm{g}}}

\def\vs{{\bm{s}}}

\DeclareMathAlphabet{\mathsfit}{\encodingdefault}{\sfdefault}{m}{sl}
\SetMathAlphabet{\mathsfit}{bold}{\encodingdefault}{\sfdefault}{bx}{n}

\DeclareMathOperator*{\argmax}{arg\,max}

\definecolor{orange(ryb)}{rgb}{0.98, 0.6, 0.01}
\definecolor{officegreen}{rgb}{0.0, 0.6, 0.0}
\title{\dataset: A Benchmark for Compositional Generalization\\on Multimodal Human Activities}

\author{Terry Yue Zhuo$^1$ \and \bf{Yaqing Liao}$^2$ \and \bf{Yuecheng Lei}$^2$ \\
         \bf{Lizhen Qu}$^1$\textsuperscript{*} \and \bf{Gerard de Melo}$^3$ \\
         \bf{Xiaojun Chang}$^4$ \and \bf{Yazhou Ren}$^2$ \and \bf{Zenglin Xu}$^{5,6}$\textsuperscript{*}\\
         $^1$Monash University  $^2$University of Electronic Science and Technology of China\\
         $^3$HPI/Univerisity of Potsdam  $^4$University of Technology Sydney \\
         $^5$Harbin Institute of Technology, Shenzhen
        $^6$Peng Cheng Lab\\
         }

\begin{document}
\maketitle
\def\thefootnote{*}\footnotetext{Corresponding authors: \texttt{lizhen.qu@monash.edu}, \texttt{xuzenglin@hit.edu.cn}}\def\thefootnote{\arabic{footnote}}

\begin{abstract}
We introduce \dataset, a novel vision-language benchmark for human activity planning. It is designed for a task where embodied AI agents can reason and forecast future actions of humans based on video clips about their initial activities and intents in text. The dataset consists of 2.9k videos from \charades extended with intents via crowdsourcing, a multi-choice question test set, and four strong baselines. One of the baselines implements a neurosymbolic approach based on a multi-modal knowledge base (MKB), while the other ones are deep generative models adapted from recent state-of-the-art (SOTA) methods. According to our extensive experiments, the key challenges are compositional generalization and effective use of information from both modalities\footnote{Our benchmark is available at \url{https://github.com/terryyz/ViLPAct}}.
\end{abstract}
\section{Introduction}
\begin{figure}[ht]
    \includegraphics[width=\columnwidth]{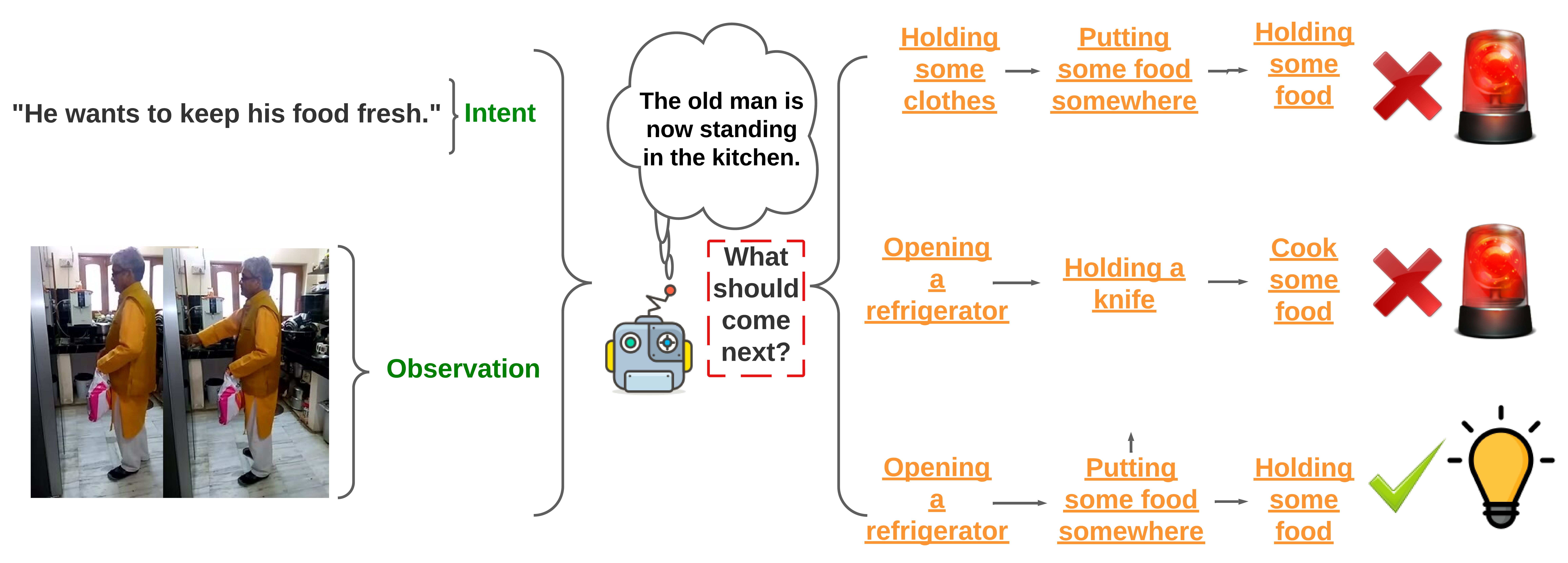}
    \caption{In daily life scenarios, an agent should be aware of future actions that will likely be taken by the user based on what it has observed. In this example, inputs of intent and observation are colored in \textcolor{officegreen}{green}, while potential future action sequences are highlighted in \textcolor{orange(ryb)}{orange}. The first two sequences contain actions which do not align with the human intent. Thus, the agent needs to automatically detect which future actions are plausible by understanding the user's intent.}
    \label{fig:Intro}
    \vspace{-2mm}
\end{figure}

One of the ultimate goals of Artificial Intelligence is to build intelligent agents capable of accurately understanding humans’ actions and intents, so that they can better serve us~\cite{kong2018humanActionRecogntionPrediction,zhuo2023exploring}. Newly emerging applications in robotics and multi-modal planning, such as Amazon Astro, have demonstrated a strong need to understand human behavior in multimodal environments. On the one hand, such an agent, e.g.\ an elderly care service bot, needs to understand human activities and anticipate human behaviors based on users' intents. Here the intents may be estimated based on previous activities or articulated verbally by users. The anticipated behaviors may be used for risk assessment (e.g.\ falling of elderly people) and to facilitate collaboration with humans. On the other hand, recent advances in robotics show that it is possible to let robots learn new tasks directly from observed human behavior without robot demonstrations~\cite{yu2018oneShot,sharma2019third}. However, that line of work focuses on imitating observed human actions without anticipating future activities.

To promote research on action forecasting based on intents, we propose the \textit{vision-language planning} task for human behaviors. As shown in Fig.~\ref{fig:Intro}, given an intent in textual form and a short video clip, an agent anticipates which actions a human is likely to take. We consider intents as given because there is already ample research on intent identification~\cite{pandey2020pedestrian} and automatic speech recognition~\cite{malik2021asrSurvey}. To the best of our knowledge, there is no dataset to evaluate models for this task. 

The task poses two major challenges. First, there are often multiple plausible action sequences satisfying an intent. Second, it is highly unlikely that a training dataset can cover all possible combinations of actions for a given intent. Hence, models need to acquire \textit{compositional generalization}~\cite{fodor1988connectionism}, the capability to generalize to unseen action sequences composed of known actions.

In this work, we construct a dataset called \dataset for \underline{Vi}sion-\underline{L}anguage \underline{P}lanning of human \underline{Act}ivities, which  to the best of our knowledge is the \textit{first} dataset studying the above challenges. Specifically, we extend the \charades dataset~\cite{sigurdsson2016hollywood} with intents via crowd-sourcing. As it is practically infeasible to find all possible future action sequences given an intent and a video clip of initial activities, we propose to evaluate all systems by letting each of them answer multi-choice comprehension questions (MQA) \textit{without training them on those questions}. Given an intent and a video clip showing initial activities, each multi-choice question provides a fixed number of future action sequences as possible answers. A system is then asked to select the most plausible action sequence among them. We show that the rankings of all models using the MQAs correlate strongly with those obtained by asking human assessors to directly observe estimated action sequences. For training, we provide both a dataset for end-to-end training of sequence forecasting and a multimodal knowledge base (MKB) built from that dataset, which is also the \textit{first} video-based multimodal knowledge base for human activities to the best of our knowledge. 


We conduct the first empirical study to investigate compositional generalization for the target task. As baselines, we adapt three strong end-to-end deep generative models for this task and propose a neurosymbolic planning baseline using the MKB. The model is neurosymbolic because it combines both deep neural networks and symbolic reasoning~\cite{garcez2020neurosymbolic}. Given a video of initial activities and an intent, the deep models generate the top-$k$ relevant action sequences, while the neurosymbolic planning model sends the intent and the action sequence recognized from the video as the query to the MKB, followed by retrieving the top-$k$ relevant action sequences. Each model selects the most plausible answers by performing probabilistic reasoning over the relevant action sequences. We conduct extensive experiments and obtain the following key experimental results:
\begin{itemize}[noitemsep,nolistsep]
    \item We compare the evaluation results using MQA with the ones of human evaluation. The results of both methods are well aligned. Thus, MQA is reliable without requiring human effort.
    \item The likelihood functions of the deep generative models are not able to reliably infer which answers are plausible. In contrast, probabilistic reasoning is an effective method to improve compositional generalization.
    \item Despite information from both modalities being useful and complementary, all baselines heavily rely on intents in textual form but fail to effectively exploit visual information from video clips. 
\end{itemize}
\section{Related Work}
\paragraph{Vision-Language Planning Task}
Vision Language Navigation (VLN) was among the first widely used goal-oriented vision-language tasks, requiring AI agents to navigate in an environment without interaction by reasoning on the given instruction \cite{anderson2018vision, hermann2020learning, misra2018mapping, jain2019stay}. Recently, further goal-oriented vision-language tasks have been proposed. The Vision and Dialogue History Navigation (VDHN) task \cite{de2018talk, nguyen2019help, thomason2020vision}, which is similar to VLN, requires agents to reason on the instructions over multiple time steps. Other tasks such as Embodied Question Answering (EQA; \citealt{das2018embodied, wijmans2019embodied}), Embodied Object Referral (EOR; \citealt{qi2020reverie, chen2019touchdown}) and Embodied Goal-directed Manipulation (EGM; \citealt{shridhar2020alfred, kim2020arramon, suhr2019executing}) rely on reasoning and interpreting the instruction with observation or object interaction in the environment. However, we argue that there are other ways to learn to plan without practising. Our task is one example of this, requiring agents to reason over the observation without performing actions.

\paragraph{Vision-Language Planning Datasets}
As existing vision-language planning datasets emphasize teaching embodied AI to perform the task like humans, they are constructed with interactive AI in mind. VLN \cite{anderson2018vision} datasets initially started exploring planning tasks with the textual instruction as a step-by-step abstract guide and minimal interaction with the environment. Extending the VLN task, VDHN \cite{de2018talk} datasets provide an interactive textual dialogue between the speaker and the receiver in multiple steps. The EQA \cite{das2018embodied} task takes this a step further by providing data in an object-centric QA manner, advancing systems to understand the given environment through object retrieval. The EOR \cite{qi2020reverie} task designs object-centric datasets with detailed instructions, aiming at localizing the relevant objects accurately.
The closest benchmark to ours is ALFRED \cite{shridhar2020alfworld} from the EGM task, which lets embodied agents decide on actions and objects to be manipulated based on detailed instructions. However, in our setting, we ask intelligent systems to predict the most reasonable future action sequence based on human intents and answers in a Multiple Choice Question Answering (MQA) format. During prediction, we still give systems the flexibility to consider various combinations of actions and objects.

\paragraph{Vision-Language Planning Modeling}
According to \newcite{francis2021core}, several approaches have been used for planning. Greedy search in end-to-end models has been reported in several studies to work well in goal-oriented tasks \cite{fried2018speaker, das2018embodied, shridhar2020alfred, anderson2018vision}. Task progress monitoring \cite{ma2019regretful} is another method to tackle the planning. It allows models to backtrack on actions if the current action is found to be suboptimal. Mapping \cite{anderson2019chasing} has as well been proposed for efficient planning via sensors. Topological and Exploration planning \cite{deng2020evolving, ke2019tactical} enables modeling the planning in a symbolic manner. When goals are provided as several sub-goals, a divide and conquer strategy \cite{misra2018mapping, shridhar2020alfred,suhr2019executing} may be invoked to perform sub-task planning. In our work, we highlight another potential approach, knowledge base retrieval. As we construct an MKB containing various action sequences with detailed features, intelligent agents can retrieve the most suitable sequence from the MKB source in order to perform the planning.
\section{Dataset Construction}

We adopt videos from \charades~\cite{sigurdsson2016hollywood} and solicit intents for videos via crowd-sourcing. We consider videos that have action sequences of sufficient length appearing in both initial video clips and answers, which result in  a dataset comprising 2,912 videos. The dataset is split into training/validation/test sets with a ratio of 70\%, 10\%, 20\%. On the training dataset, we build an MKB by incorporating structural and conceptual information. On the test dataset, we collect a set of MQAs for model evaluation. The evaluation with MQAs is in fact an adversarial testing method, widely used for quality estimation in machine translation~\cite{kanojia2021pushing}. Herein, the ability of a model to discriminate between correct outputs and meaning-changing perturbations
is predictive of its \textit{overall} performance, not just its robustness. Thus MQAs are applied only for testing. 

\subsection{Data Normalization and Filtering}
\charades is a large-scale video dataset of daily indoors activities collected via Amazon Mechanic Turk\footnote{\url{https://www.mturk.com}} (AMT). The average length of videos is approximately 30 seconds. It involves interactions with 46 object classes and contains 157 action classes, which are also referred to as \textbf{actions} for short. Each action is represented as a verb phrase, such as ``pouring into a cup".  This dataset is chosen because \textit{i)} it contains a sufficient number of long action sequences of human daily activities; \textit{ii)} the intents are easily identifiable, as the activities in the videos are based on scripts; \textit{iii)} there are rich annotations of videos that can be leveraged for dataset construction. The details of action sequence selection in videos are presented in Appendix \ref{algo:action_seq}, with the goal of choosing core action sequences having clear human goals.

In order to assess the quality of extracted action sequences, we randomly sample 100 videos from the test set for manual inspection. The primary action sequence of each video is evaluated in terms of three criteria: \textit{i)} if all actions of a sequence occur in the video; \textit{ii)} if the actions of a sequence appear in the same order as in the video; \textit{iii)} if a sequence has any actions missing between the first and the last action. In total, we determined that 94 videos have all actions of their action sequences covered in the video. The actions of 92 videos appear in the same order as in the videos. Furthermore, 85 videos have no actions missing between the first and the last action of their sequences. Thus, the quality of such action sequences is adequate for VL planning evaluation.

Following prior work~\cite{ng2020forecasting}, we consider the first 20\% of a video as its initial visual state and aim to forecast future actions appearing in the remaining part of the video for a given intent. To have at least one future action per video, we retain only videos that contain at least one action sequence comprising more than three actions. As a result, we obtain 2,912 such videos, each of which is associated with one action sequence of length longer than three. 

\subsection{Intent Annotation}
An \textit{intent} may be defined as ``something that you want and plan to do''.\footnote{Cambridge Dictionary, \url{https://dictionary.cambridge.org/}} Philosophers distinguish between future-directed intents and present-directed ones~\cite{cohen1990intention}. The former guide the planning of actions, while the latter causally produce behavior. As the focus of this work is anticipating and planning actions, we encourage crowd-workers to also provide future-directed intents. 

We recruit crowd-workers to annotate videos with future-directed and present-directed intents. Each annotator is provided with a full video clip and the associated action sequence. They are instructed to answer the question \textit{what the person wants to do by taking the actions in the video}. Every annotator is asked to submit two intents. One of them should describe which activity the person intends to take, such as ``drink a glass of water''. The other one needs to be at a high-level, such as ``quench the thirst'' or ``be thirsty''. The permitted formats are either ``S/He wants to $+$ \textit{do\_something}'' or ``S/He is $+$ \textit{feeling}". Thus, the annotators are encouraged to provide future-directed intents by differentiating them from ones causally leading to behaviours. To ensure the quality of intent annotations, we randomly assign three crowd-workers to write intents per video. The process of constructing the dataset for intent annotation involved a rigorous validation and selection process. One of the authors acted as an expert annotator, and conducted a thorough review of all crowd-sourced intents to identify and select the most reasonable annotations as the final results. The validation process was completed in three rounds, yielding increasingly higher percentages of reasonable annotations, with 82\%, 94\% and 100\% respectively for each round. The annotations that did not meet the required criteria were discarded and not included in the final dataset. This rigorous validation process ensured that the final dataset is comprised of high-quality and relevant annotations, providing a robust foundation for subsequent modeling and analysis.

\subsection{Multimodal Knowledge Base}
\label{sec:mkb}
\label{MKB}
We construct the MKB of human activities based on the \textbf{training set} and \textbf{validation set} by taking a neurosymbolic approach. The main challenges herein are twofold: i) how to represent multimodal information from videos, action names, and intents adequately to facilitate information retrieval; ii) how to model shared knowledge of multimodal information. For the former, we allow both string and embedding based retrieval methods by attaching neural representations of video clips and texts to symbols of actions and action sequences. For the latter, we employ the classical planning language STRIPS~\cite{bylander1994STRIPS} and neural prototypes to encode abstract properties of actions.

At the core of the MKB is a knowledge graph $\mathcal{G} = (\mathcal{V}, \mathcal{E})$, where the node set $\mathcal{V}$ comprises four types of nodes: action classes, action video clips, action sequences, and action sequence videos, while the edge set $\mathcal{E}$ contains edges reflecting relationships between nodes. 

An \textit{action class} $a^c$ is the abstraction of an action described in the language of STRIPS. The attributes of an action class include its ID, its name $\tau$, its precondition set \textsc{PRE}, its add effect set \textsc{ADD}, and its delete effect set \textsc{DEL}. An action is executed only if its preconditions are satisfied. The effect sets \textsc{ADD} and \textsc{DEL} of an action class describe the add and delete operations applied to the current state after executing the action. For example, the precondition of \act{Closing a refrigerator} is \act{isOpen(refrigerator)}, \act{ADD = isClosed(refrigerator)} and \act{DEL=isOpen(refrigerator)}. In this way, the properties described in STRIPS present the shared knowledge of each action class. 

\begin{figure}[h]
    \centering
    \includegraphics[width=\linewidth]{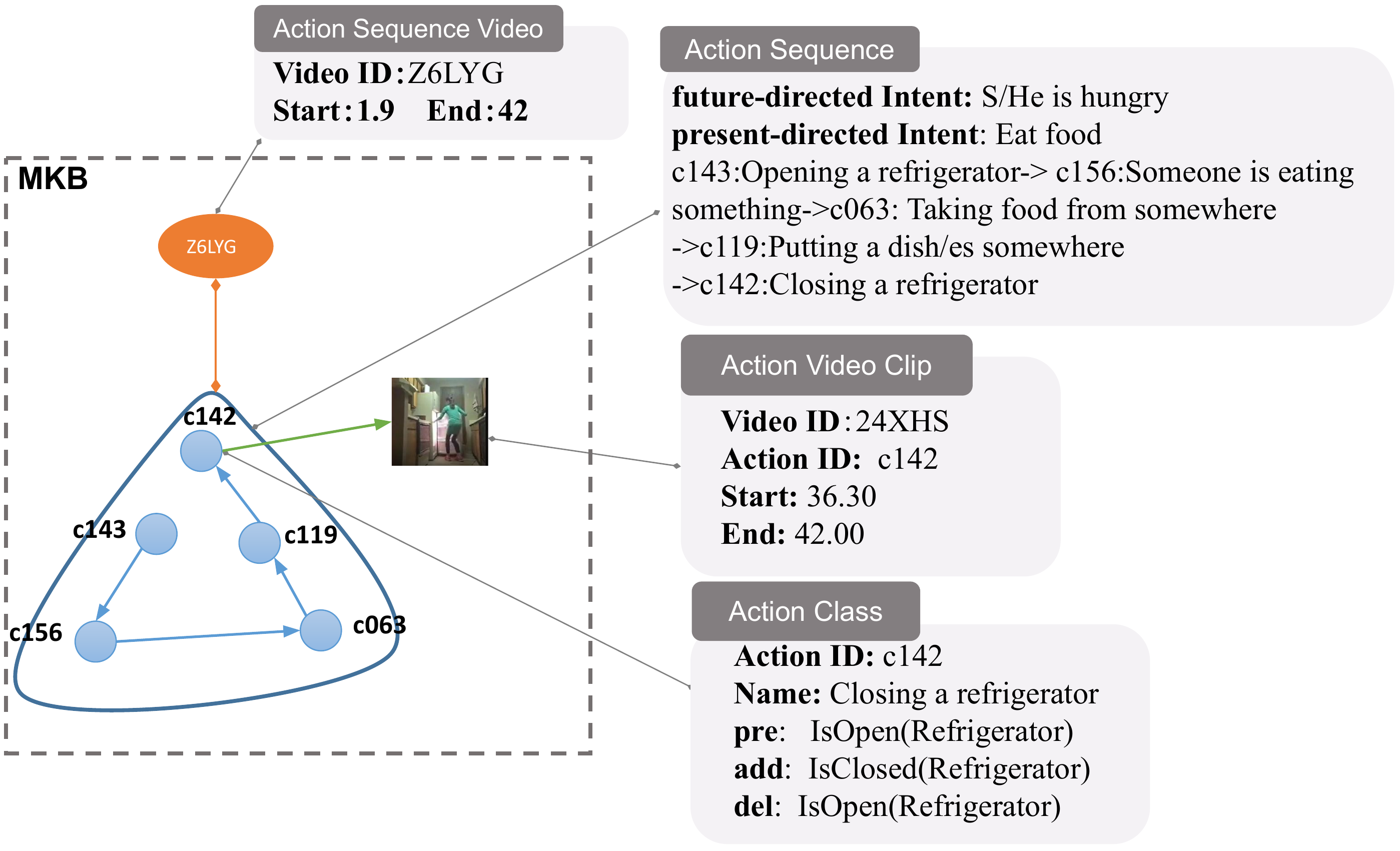}
    \caption{An example action sequence in the MKB.}
    \label{fig:MKB}
    \vspace{-1ex}
\end{figure}
 
An \textit{action sequence} comprises a future-directed intent, a present-directed intent, and a sequence of action IDs. An intent is represented by both a word sequence and the distributed representation of the word sequence. We obtain the distributed representation of an intent by applying BERT~\cite{devlin2018bert} and utilizing the representation of the CLS token. The collection of action sequences can be easily turned into a training set for end-to-end models by associating them with the corresponding video files.
    
    
The MKB includes two types of visual nodes: \textit{action sequence videos} and \textit{action video clips}. Each action sequence video is linked to the corresponding action sequence. For each action in an action sequence, we associate it with the corresponding video clip, as illustrated in Fig.~\ref{fig:MKB}. For each action video clip, we apply \itd to encode it into a sequence of frame-level visual feature vectors $\{\mathbf{f}_{s_1}, \mathbf{f}_{s_2}, \dots, \mathbf{f}_{s_t}\}$, where each vector $\mathbf{f}_{s_i}\in \mathbb{R}^{1024}$ corresponds to the features of an 8-frames snippet. To represent an action sequence video, we apply average pooling to the distributed representations of all involved video clips. 

\paragraph{Relations.} We consider two types of relations in the MKB. The first type of relation links an action sequence to the corresponding visual representation. The other type of relation associates an action in an action sequence with the corresponding action class. Therefore, it is easy to perform symbolic reasoning by using the STRIPS properties of each action class involved in an action sequence.

\paragraph{Statistics of MKB} Table \ref{table:Statistics of KB} provides statistics of the MKB. As we can observe, the MKB contains 2,402 action sequence videos and 12,118 action video clips. Each action sequence video is associated with one corresponding action sequence. There are 157 action classes in total and 1,969 unique action sequences. The average length of action sequences is 5.04. 

\begin{table}[!htb]
\centering
\vspace{-1ex}
\begin{adjustbox}{width=0.8\columnwidth}
\begin{tabular}{lc}
\toprule
Item & Statistics \\
\midrule
\# of action classes & 157 \\
\# of action sequence videos & 2,402 \\
\# of action video clips & 12,118 \\
\# of action sequences (distinct seq) & 2,402 (1,969) \\
\# of action state templates & 32\\ 
\# avg. \# of action sequence length & 5.04\\\bottomrule
\end{tabular}
\end{adjustbox}
\caption{Statistics of the MKB / training + validation set}
\label{table:Statistics of KB}
\vspace{-5mm}
\end{table}
\FloatBarrier

\subsection{Multi-Choice Comprehension Questions for Evaluation}
\label{sec:multichoice}
\begin{figure}[htb]
    \centering
    \includegraphics[width=1\linewidth,trim=1.9cm 0cm 1.9cm 0cm]{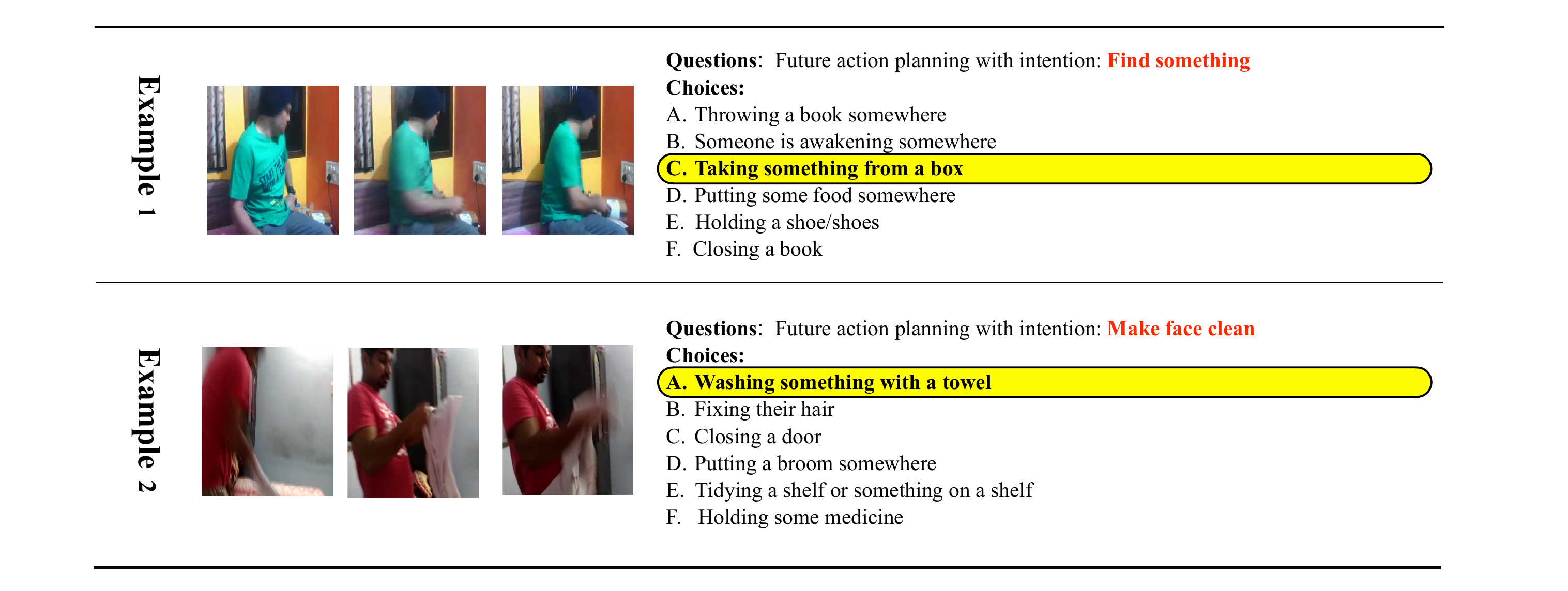}
    \caption{Two examples of \dataset MQA task}
    \label{fig:examples_1}
    \vspace{-3mm}
\end{figure}

Given the first 20\% of a video as the initial state $\vs$ and a future-directed intent $\vg$ in text, the planning evaluation task involves choosing the most plausible future action sequence $\va^f$ among six available choices. We determine the initial action sequence $\va^i$ by checking if an action of a sequence starts before the end time of the initial state. To build such a dataset, we extended the test set with adversarially generated incorrect answers. As the automatic approach may generate reasonable action sequences, we recruit another group of students to manually check all answers and determine the most plausible ones as the correct answers on AMT. Figure~\ref{fig:examples_1} shows an example of our planning task.

\paragraph{Generation of Incorrect Answers.} We adapt the \textit{Adversarial Matching} (AM) algorithm \cite{zellers2019recognition} to turn the action sequence generation task into a multi-choice test. The key idea here is to substitute an action of an observed action sequence for an alternative action that is relevant to the preceding actions and is not overly similar to the action to be replaced. As many videos in the test set have only a single future action, the AM algorithm is extended to optionally insert a future action to generate an answer candidate.

More specifically, given the initial state, the action sequence, and the intent $(\vs, \va, \vg)$ of a video, where $\va = (\va^i, \va^f)$, the algorithm starts by randomly deciding if it applies substitution or insertion to generate an answer candidate. If insertion is chosen, it inserts an action randomly selected among the 157 candidate actions, at a position that is randomly picked after the last action in $\va^i$. If instead substitution is chosen, we feed the initial action sequence $\va^i$ to \bert and use the representation of the CLS token as the representation of $\va^i$. Then we apply \bert to turn each action into a vector by using the corresponding CLS representation. We randomly pick a future action $a_i$ in $\va^f$ and compute the score of a candidate action $a_j$ as
\begin{equation}
\vspace{-0.5mm}
\begin{split}
s(a_j)=& \log(P_\mathrm{sim}(\va^i,a_j))\\
& +\lambda \log(1-P_\mathrm{sim}(a_i,a_j)),
\vspace{-0.5mm}
\end{split}
\end{equation}
where $P_\mathrm{sim}()$ is defined as cosine similarity. We set $\lambda = 0.7$ to find an optimal tradeoff between the obfuscation level of an incorrect answer and the probability of being a reasonable answer. We repeat this process until we have generated five answer candidates. For each set of generated answer candidates, we manually checked the grammaticality and fixed all the errors.

\paragraph{Quality Check via Crowd-Sourcing.} We hired three crowd-workers per video on AMT to ascertain the quality of all auto-generated answers. For each video, a worker is presented with the first 20\% of the video and the future-directed intents, which are paired with six answer candidates each (an original action sequence and five generated ones), because there were two annotators working on each video. They were instructed to choose the most reasonable pair of intent and action sequence among all possible combinations.


After checking the answers of all questions in the \textbf{test set}, we apply a set of heuristic rules to determine the final answer to each question. 
We calculate inter-annotator agreement by asking the group of workers that did the annotation to work on a sample of multi-choice questions of the MQA task. To evaluate the quality of the MQA choices, we determined the number of agreements between the ground truth (the correct answers) and the predicted answers. Then, we computed the number of agreements that would be expected by chance based on the distribution of answers. The corresponding Cohen's kappa coefficient \cite{kraemer2014kappa} is 0.91, which demonstrates the high quality. 


\begin{table}[]
\begin{adjustbox}{width=\columnwidth}
\begin{tabular}{lc}
\toprule
Statistics & Value \\ \midrule
\# of videos & 510 \\
    avg. \# of observed actions & 2.79 \\
    avg. \# of future actions & 2.40 \\
    avg. \# of actions  & 5.19 \\
    \# of full action seq occurring in the training set &121 \\
    avg. \# of distinct future action sequences for an intent & 2.16\\
    std.\ dev.\ of \# of distinct future action sequences for all intents & 3.69\\
\bottomrule
\end{tabular}
\end{adjustbox}
\caption{Basic statistics of MQA task / test set}
\label{table:Statistics of tasks}
\vspace{-5mm}
\end{table}

Table \ref{table:Statistics of tasks} shows the basic statistics of the test set. The average number of observed actions in $\vs$ is similar to the average number of future actions. Although all actions in the test appear in the training set, the most plausible action sequences of almost 400 videos are unseen in the training set. For intents in MQA, we also calculate the number of distinct future action sequences for each of them, and the standard deviation across all of them. The results indicate how diverse potential future action sequences can be for a single intent. Other details of MQA can be found in Appendix \ref{appendix:data_details}.

\section{Baselines}

VL plannning of human activities requires predicting future action sequences given an initial visual state video and an intent provided in textual form. The task poses two major challenges. First, information provided in two modalities are complementary to each other, while the majority of multimodal research focuses on the shared information by exploring fusion techniques~\cite{guo2019deepMultimodalSurvey}. Second, the output space is exponentially large with respect to the action space. It is not realistic to assume that all action sequences are already observed in the training data. Hence, any models to tackle this task are expected to address \textit{systematic composition}~\cite{fodor1988connectionism} of human activities, the capacity to understand and produce a huge number of novel combinations of known actions. In contrast, state-of-the-art deep learning methods often perform poorly on compositional generalization~\cite{lake2019compositional,keysers2019measuringCompositional}. 

We compare deep generative models and a neurosymbolic planning model in the framework of retrieval and reasoning. Given the first 20\% of a video and a future-directed intent, the first step is to obtain top-$k$ relevant action sequences, followed by performing reasoning over the top-$k$ action sequences to find the most plausible answers. Both types of models share the same reasoning module but differ in how they obtain top-$k$ action sequences. For reproducibility, the details of all models are provided in Appendix \ref{appendix:deep_generative} and \ref{appendix:neurosymbolic}.


\subsection{Deep Generative Models}
The deep generative models apply beam search to produce the top-$k$ most likely future action sequences, followed by performing reasoning.


\paragraph{\uniVL} We adapt UNIVL~\cite{luo2020univl} for the target task (denoted as \uniVL), which is a SOTA unified pretrained vision-language model for multimodal understanding and generation. We consider \uniVL because it performs the best on the tasks that are closest to our target task, such as YouCook2 \cite{zhou2017automatic}. The pre-trained \uniVL takes as input an intent and an initial video clip, and is fine-tuned to forecast future action sequences.


\paragraph{Two Stage Planning Model.} The two stage planning baseline, \textbf{TwoStagePlan} for short, starts by converting an initial video clip into an action sequence in text by using \uniVL, followed by applying a pre-trained language model, ProphetNet \cite{qi2020prophetnet} (denoted as \prophetNet for \dataset), to predict future actions. 



\paragraph{\prophetNet} To study the impact of visual information, we consider a text-only baseline by employing \prophetNet to predict future action sequences only based on intents.

\subsection{Neurosymbolic Planning Model}

Given an intent and an initial visual state, the neurosymbolic planning model (\textbf{NSPlan}) retrieves top-$k$ relevant action sequences from the MKB in two stages, and then utilizes the retrieved results to infer the most plausible answers. 
In the first stage, we apply the pretrained \uniVL to convert a video clip into an action sequence and send it as a query to the MKB to retrieve top-$50$ results. For each retrieved result, the ranking score is the weighted sum of the \bmTF~\cite{robertson1994bm25} score between two action sequences and the cosine similarity between the intents.

In the second stage, it re-ranks the initial retrieval results by using both visual and symbolic knowledge. Each retrieved action sequence is represented as a sequence of frame-level visual feature vectors, extracted by the visual encoder \itd. An Ordered Temporal Alignment Module (OTAM; \citealt{cao2020few}) is applied to compare two visual feature sequences. In order to rank the sequences with potential future actions higher, we use a rule-based score function to prefer longer sequences containing unseen actions. In the end, we keep only the top-$k$ results for probabilistic reasoning.

\subsubsection{Probabilistic Reasoning for MQA}
We propose a novel approach for MQA called \probInf, which, based on the top-$K$ action sequences, performs probabilistic inference over the retrieved action sequences to identify the most likely answer for a question. 
From each retrieved result after re-ranking, obtained from \RetrievalScoring, we remove the predicted observed action sequence $s^a_q$ to obtain potential future action sequences. For generative models, we directly use the generated outcomes. For each answer candidate $c_i$ of a question, we compute $p(c_{i} \mid \vs, \vg)$ by integrating over all retrieved results $\{r_1, r_2, ..., r_K\}$, given the initial visual state $\vs$ and intent $\vg$:
\vspace{-2mm}
\begin{equation}
\label{eq:prob_inf}
p(c_{i} \mid \vs, \vg)=\sum_{k=1}^{K} p( c_{j}  \mid r_{k})\, p(r_{k} \mid \vs, \vg ),
  \vspace{-2mm}
\end{equation}
where $p(r_j \mid \vs, \vg) = \frac{\exp(s_f(r_j))}{\sum_{k=1}^K \exp(s_f(r_k))}$ is the normalized ranking score for a result $r_j$ and $p( c_{j}  \mid r_{k})$ is the normalized similarity between an answer candidate and each retrieved result. As both answers and retrieved results are action sequences represented in text, we employ the time series metric Time-warped edit distance (TWED; \citealt{TWED}) to compute their similarity as $\phi(f(c_i), f(r_j)) = 1 - d_{\text{twed}}(f(c_i), f(r_j)) / \max(|c_i|, |r_j|)$, where $f(c_i)$ denotes the visual prototype representation of an action sequence and $d_{\text{twed}}(f(c_i), f(r_j))$ denotes the distance computed by TWED algorithm. Then the normalized similarity over $n$ possible answers of a question is given by:
\begin{equation}
P\left(c_{i} \mid r_{j}\right)=\frac{\exp \phi(f(c_i), f(r_j))}{\sum_{k=1}^{n} \exp \phi(f(c_k), f(r_j))}
\label{conditional prob}
\end{equation}
The most plausible answer is the one with the maximal $p(c_{j} \mid \vs, \vg)$ over all answer candidates.

\section{Experiments}

We conduct extensive experiments to answer the following three main research questions. The other research questions are addressed in Appendix \ref{appendix:other_rqs}.

\paragraph{RQ1: How reliable is the MQA evaluation method?}
\label{rq1}

 \begin{table}[ht]
\begin{adjustbox}{width=\columnwidth}
 \begin{tabular}{|c|c|c|c|c|}
 \hline
 Method & \UniVLProphetNet & \RetrievalScoring & \prophetNet & \uniVL \\\hline
  Log-likelihood Accuracy(\%) & 19.02  & - & 10.78 & 22.35 \\\hline
  top-$1$ Reasoner-scoring Accuracy(\%)  & 63.72 & 60.58 & 67.45 & 69.01 \\\hline
 top-$10$ Reasoner-scoring Accuracy(\%)  & 60.19  & 64.11 & 69.01 & 70.58 \\\hline
 \end{tabular}
\end{adjustbox}
 \caption{Comparison of all systems, with \textbf{Human performance} of 94.25\% accuracy, which is obtained by asking humans to answer the MQAs directly.}
  \label{table:Comparison of Retrieval-based method, Generation-based method}
  \vspace{-3mm}
 \end{table}



 \begin{figure}[htbp]
\centering 

\begin{minipage}[t]{0.48\linewidth}
\centering
\includegraphics[width=\linewidth]{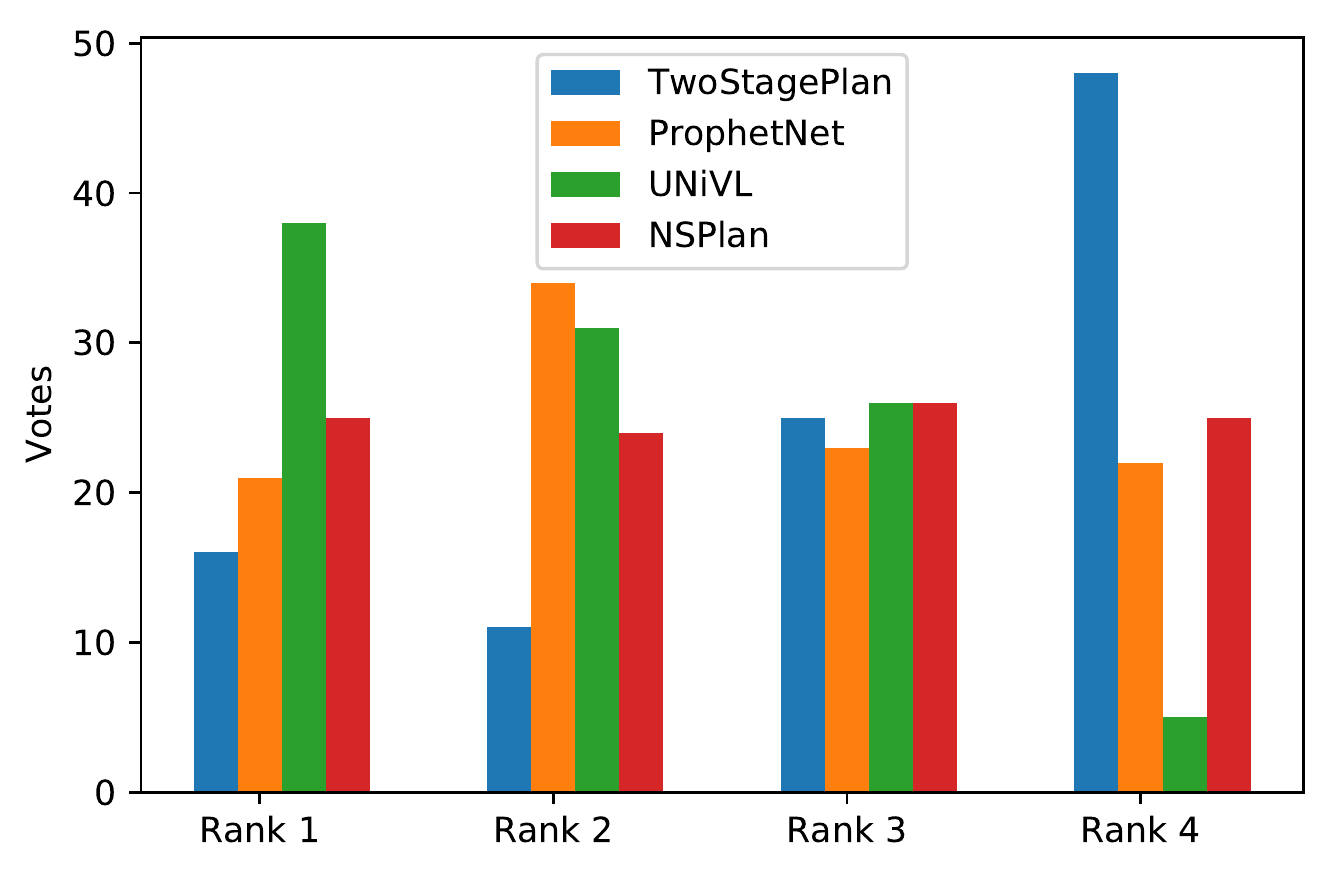}
\end{minipage}
\begin{minipage}[t]{0.48\linewidth}
\centering
\includegraphics[width=\linewidth]{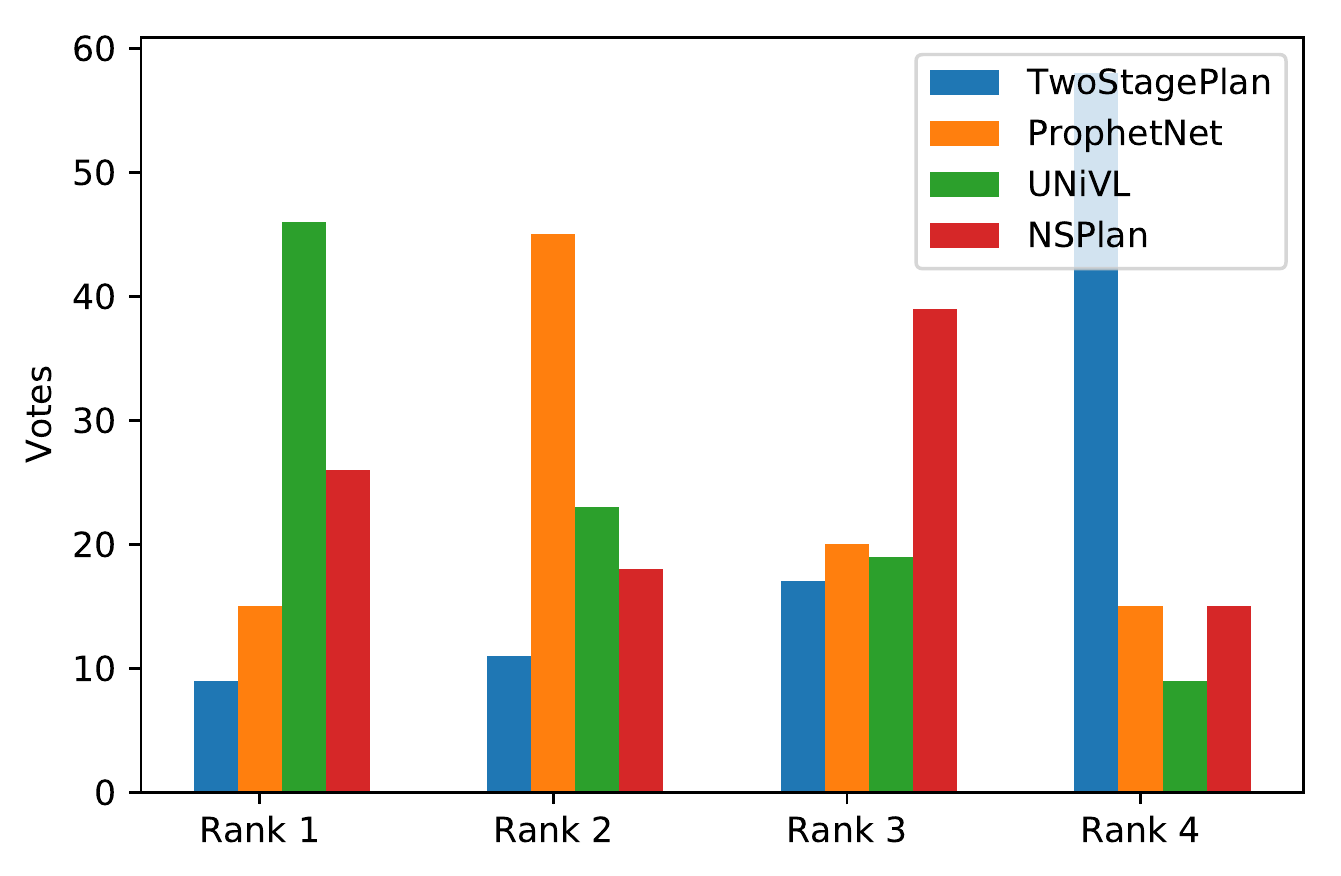}
\end{minipage}
\caption{Human evaluation on the quality of top-$10$ (left) \& top-$1$ (right) future action sequence.}
\label{fig:human_model_evaluation_topk}
  \vspace{-5mm}
\end{figure}

We show that the evaluation results using MQA are consistent with those by asking humans to directly observe model outputs. For this, we recruit five crowd-workers to rank all models in comparison on each of the 100 questions randomly sampled from the test set, and compare them with the corresponding results using MQA. Specifically, for each question, a crowd-worker is asked to rank the top-$k$ outputs of the four baselines in terms of how well they match the intent and the remaining $80\%$ of the original videos. As a result, Figure \ref{fig:human_model_evaluation_topk} shows how frequent each model is ranked at position $X$ judged by the crowd-workers w.r.t. the top-$10$ predictions (left) and top-$1$ predictions (right), respectively. In both cases, we consistently find that the best model is \uniVL, followed by \prophetNet, \RetrievalScoring, and \UniVLProphetNet. The ranking result is the same as using MQA on the same set of questions. The ranking differences on individual questions between the human evaluation and MQA are statistically insignificant according to Wilcoxon's signed-rank test~\cite{woolson2007wilcoxon}, details of which can be found in Appendix \ref{appendix:wilcoxon_test}.
 \begin{table}[h]
 \centering
 \adjustbox{max width=\columnwidth}{
 \begin{tabular}{|c|c|c|c|c|c|}
 \hline
 Method  & \UniVLProphetNet & \RetrievalScoring & \prophetNet & \uniVL \\ \hline
 Seen Accuracy(\%)  & 60.33 & 65.28 & 70.24 & 74.38 \\\hline
  UnSeen Accuracy(\%)  & 60.15  & 63.75 & 68.63 & 69.40\\\hline
 \end{tabular}}
 \caption{Top-$10$ Reasoner-scoring Accuracy on seen and unseen action sequences. Seen data refers to the MQAs with plausible action sequences observed in the training data. Unseen data refer to the ones with plausible action sequences not observed in the training data.}
 \label{table:unseen or seen}
 \vspace{-3mm}
 \end{table}


\paragraph{RQ2: What are the key challenges?} We identify two major challenges of the target task.

\paragraph{Compositional Generalization Using Reasoning.} It is common practice to rank each answer by the likelihood yielded by a generative model~\cite{holtzman2021surface}. However, Table \ref{table:Comparison of Retrieval-based method, Generation-based method}, which provides the overall evaluation results using MQA, shows that the generative baselines perform poorly when they rank answers based on the likelihood. In contrast, \probInf effectively uses top-$k$ results to boost the performance of all generative models by more than 44\%. For the respective performance on seen and unseen action sequences (Table \ref{table:unseen or seen}), \probInf delivers stable results across models. \textit{The performance on unseen combinations of seen actions measures exactly the ability of compositional generalization}. This raises the question of \textit{``Why \probInf helps compositional generalization ?''} for future research. As there is still a sizable gap between seen and unseen action sequences, and all models fall short of the human performance (Table \ref{table:Comparison of Retrieval-based method, Generation-based method}) by at least 23\%, how could we make further improvements?

\paragraph{Effective Use of Both Modalities.} To understand the utility of each modality, we compare the two strongest multimodal models by varying their inputs: including both modalities or just a single modality. As shown in Table \ref{table:ablation}, intents provide the strongest signal, while visual information is useful overall for both models. This also explains why \prophetNet comes close to \uniVL.

 \begin{table}[h]
 \centering
 \scalebox{0.7}{
 \begin{tabular}{|c|c|}
 \hline
  \uniVL w/o Vision & \uniVL \\ \hline
  69.01 & 70.58 \textcolor{red}{$\uparrow$}\\\hline
  \RetrievalScoring w/o Vision & \RetrievalScoring \\ \hline
  61.56 & 64.11\textcolor{red}{$\uparrow$} \\\hline
  \uniVL w/o Intent & \uniVL \\ \hline
 61.17 & 70.58 \textcolor{red}{$\uparrow$} \\\hline
 \RetrievalScoring w/o Intent & \RetrievalScoring \\ \hline
 60.78 & 64.11 \textcolor{red}{$\uparrow$} \\\hline
 \end{tabular}}
 \caption{Modality study on MQA accuracies (\%) of different baselines via Reasoner-scoring.}
  \label{table:ablation}
  \vspace{-3mm}
 \end{table}
 
 To further investigate the significance of visual information for multimodal models, we substitute the visual features of \uniVL for randomly selected ones during both training and inference, finding that \uniVL suffers from only a 4\% drop of accuracy using MQA. Hence, the multimodal models capture only weak associations between visual features and future action sequences.
 
 It is counter-intuitive that visual features do not play a significant role, because plans vary in accordance with different visual environments. We conjecture this is due to poor performance of action recognition. To verify this, we feed ground-truth actions observed in the first 20\% of videos to both \UniVLProphetNet and \RetrievalScoring during training and inference. They reach an accuracy of 82.11\% and  81.37\% respectively, improved by more than 15\%.

\paragraph{RQ3: To what degree can the top-$k$ results reflect the performance differences of systems?}
The reasoning method \probInf leverages the top-$k$ results produced by the models, hence it is useful to inspect those results for further insights. Therefore, we compare the top 10 results of each model in terms of precision and recall by treating each action sequence as a set~\cite{ng2020forecasting}, as well as seq-hits@5 for measuring exactly matched action sequences. 
Moreover, to investigate the \textit{diversity} of the top-$k$ lists, we consider Dist1 and Dist2 \cite{li-etal-2016-diversity}, which respectively measure the number of unique action and consecutive action pairs in the top-$k$ lists. The definitions of a complete list of used metrics and their results are provided in Appendix \ref{appendix:metrics} and \ref{appendix:future_action_sequence}. 

According to Table \ref{main:table: Comparison of top-K future sequences of two baselines}, \uniVL outperforms all other models in terms of quality-oriented metrics but falls short of \prophetNet in terms of both diversity metrics. However, none of the metrics obtains the same ranking of models in accordance with the human evaluation. Although \RetrievalScoring achieves higher recall than \prophetNet, its precision and seq-hits@5 are significantly lower than those of \prophetNet, explaining why it performs worse than \prophetNet using MQA.

\begin{table}[]
\begin{adjustbox}{width=\columnwidth}
 \begin{tabular}{|c|c|c|c|c|c|}
\hline
 \multirow{2}{*}{Setting} & \multicolumn{3}{c|}{Quality} & \multicolumn{2}{c|}{Diversity} \\ \cline{2-6} 
\multicolumn{1}{|c|}{} & precision & \multicolumn{1}{c|}{recall} & \multicolumn{1}{c|}{seq-hits@5} & \multicolumn{1}{c|}{Dist1} & Dist2 \\ 
\hline
\UniVLProphetNet &21.59 &15.59 &10.00 &32.55&58.54 \\\hline
\RetrievalScoring
&20.73&21.66&5.69&38.46&66.34\\\hline
\prophetNet
&21.35 &9.61 &8.12 & \textbf{51.96} &\textbf{81.93}\\\hline
\uniVL
&\textbf{23.67} &\textbf{22.02} &\textbf{12.75}  &47.10 &77.42\\\hline
\end{tabular}
\end{adjustbox}
\caption{Comparison of top-10 future sequences}
\label{main:table: Comparison of top-K future sequences of two baselines}
\vspace{-4mm}
\end{table}

\section{Conclusion}

We construct the novel benchmark \dataset to evaluate the ability of systems to anticipate and plan human actions in a multimodal vision-language setting, with a focus on evaluating their compositional generalization capabilities. In this benchmark, we extend \charades with intents, construct a test set with multi-choice questions, and include four strong baselines. Our empirical studies demonstrate that the task is easy for humans, but challenging for SOTA deep learning models due to the need for compositional generalization and an effective use of information from both modalities. The neurosymbolic planning baseline shows a promising research avenue for using symbolic and multimodal knowledge in an MKB. 

\section*{Ethical Considerations}

In order to mitigate the potential for exposure to problematic content in the Charades video dataset, we have implemented stringent safety measures to safeguard our annotators against adverse psychological effects. To ensure the suitability of the video content, the authors initially conducted a comprehensive review. However, it is recognized that the process of annotating feedback may still result in the exposure to potentially disturbing or offensive material. To mitigate this, we only engage annotators who are of legal age and clearly communicate that discretion is strongly advised when engaging in the annotation process. In the event that an annotator experiences discomfort or distress, we provide information on how they can seek support from the Substance Abuse and Mental Health Services Administration (SAMHSA)\footnote{\url{https://www.samhsa.gov/}}, a free and confidential resource available 24/7. In addition, we have established a feedback mechanism to allow annotators to communicate their concerns in real-time. Our response time to any feedback received is within 24 hours. Furthermore, we compensate our annotators with competitive wages, with an average hourly rate of approximately \$12.

\section*{Limitations}
In this work, we have proposed a new vision-language benchmark for compositional generalization on human activities. Although it contains numerous videos and diverse actions, it only emphasizes in-door activities, which is a subdomain of human activities. We encourage future research to investigate the compositional generalization on various scenarios of outdoor activities. In addition, despite the fact that our benchmark contains a reasonable number of actions, these actions are constrained by limited types of verb and noun phrases, due to the nature of \charades. We suggest the development of a more extensive dataset covering open-vocabulary actions in future applications.

\section*{Acknowledgements}
This work was partially supported by
the National Key Research and Development Program of
China (No. 2018AAA0100204), a key program of fundamental research from Shenzhen Science and Technology
Innovation Commission (No. JCYJ20200109113403826), the Major Key Project of PCL (No. PCL2021A06), an Open Research Project of Zhejiang Lab (NO.2022RC0AB04),  and Guangdong Provincial Key Laboratory of Novel Security Intelligence Technologies (No. 2022B1212010005).

\bibliography{anthology}
\bibliographystyle{acl_natbib}
\section{Appendix}
\subsection{Action Sequence Extraction Algorithm}
\label{algo:action_seq}
Each video of \charades is annotated with actions from at least one action sequence. The starting and ending points of an action are labelled, but it is not clear which actions jointly meet an intent. Therefore, we implement the greedy method in Algorithm~\ref{act_algo} to automatically extract action sequences with clear intents from videos. For each video, the algorithm aims to identify a sequence of temporally and semantically coherent actions, which interact with the same or related objects. The scoring functions in Algorithm~\ref{act_algo} measure coherence from three perspectives: i) semantic relevance based on TF-IDF \cite{tfidf} reweighted Word2Vec embeddings~\cite{word2vec}, ii) temporal relevance, iii) task relevance. Each action is assigned to one of 22 tasks manually, for example, "Opening a book" and "Closing a book" are assigned to the same task.   

\begin{algorithm}
\small
\SetAlgoNoLine
\caption{Extract Action Sequences}
  \label{act_algo}  
  \KwIn{$Actions = \{a_1,a_2,\ldots ,a_n\}$, each action  $a_i = \left < cls^{a_i}, t_s^{a_i}, t_e^{a_i}\right >$, where $cls^{a_i}$ is the action class, $t_s^{a_i}$ and $t_e^{a_i}$ is the start time and end time of action $a_i$. Relevance threshold }
  \KwOut{$Activities = \{A_1, A_2, \dots ,A_n\}$, where each activity represents an action sequence}
  Remaining actions set $R_a$ = $Actions$\\
  \While{$R_a \neq \emptyset$}
  {
        Sort $R_a$ in ascending order by start time $t_s$\\
        pre action $a = R_a[0]$\\
        Activity $A = \{a\}$\\
        $Search = True$\\
        \While{Search}{
            candidates $C_a = \{a_j \in R_a | t_s^{a_j} \ge t_s^{a}\}$\\
            \For{$a_j \in C_a$} 
            {
            Calculate relevance score: $s_{a_j} = \operatorname{score}\left(a, a_{j}\right)=f_{\text {semantic }}\left(a, a_{j}\right)+f_{\text {time}}\left(a, a_{j}\right)+f_{\text {task}}\left(a, a_{j}\right).$\\
              
            Where $f_{\text {semantic }}\left(a, a_{j}\right)=\operatorname{cosine}\left(E_a, E_{a_j}\right),$\\
            $E_{a} = \sum\limits_{w \in cls^{a}} TFIDF\left(w \right) * w2v\left(w \right),$\\
            $f_{\text {time }}\left(a, a_{j}\right)=(1 - \mathrm{atanh}(|t_s^a - t_s^{a_j}|) * \pi / 2)$\\

            $f_{\text{topic}}\left(a, a_{j}\right) = \mathbf{1}\left(task^a  = task^{a_j}\right)$
         }
         $a_{\max} = \mathrm{argmax}(\{s_{a_j}| a_j \in C_a\})$\\
         \eIf{$s_{a_{\max}}$ < threshold (1.3 by optimization)}{
            Append $A$ to $Activities$\\
            Search = False
         }
         {
            Add $a_{\max}$ to Activity\\
            Remove $a_{\max}$ from $R_a$\\
            pre action $a$ = $a_{\max}$
         }
         }
  }
\end{algorithm}
\subsection{Other Data Details}
\label{appendix:data_details}
\begin{figure}[H]
    \includegraphics[width=\linewidth]{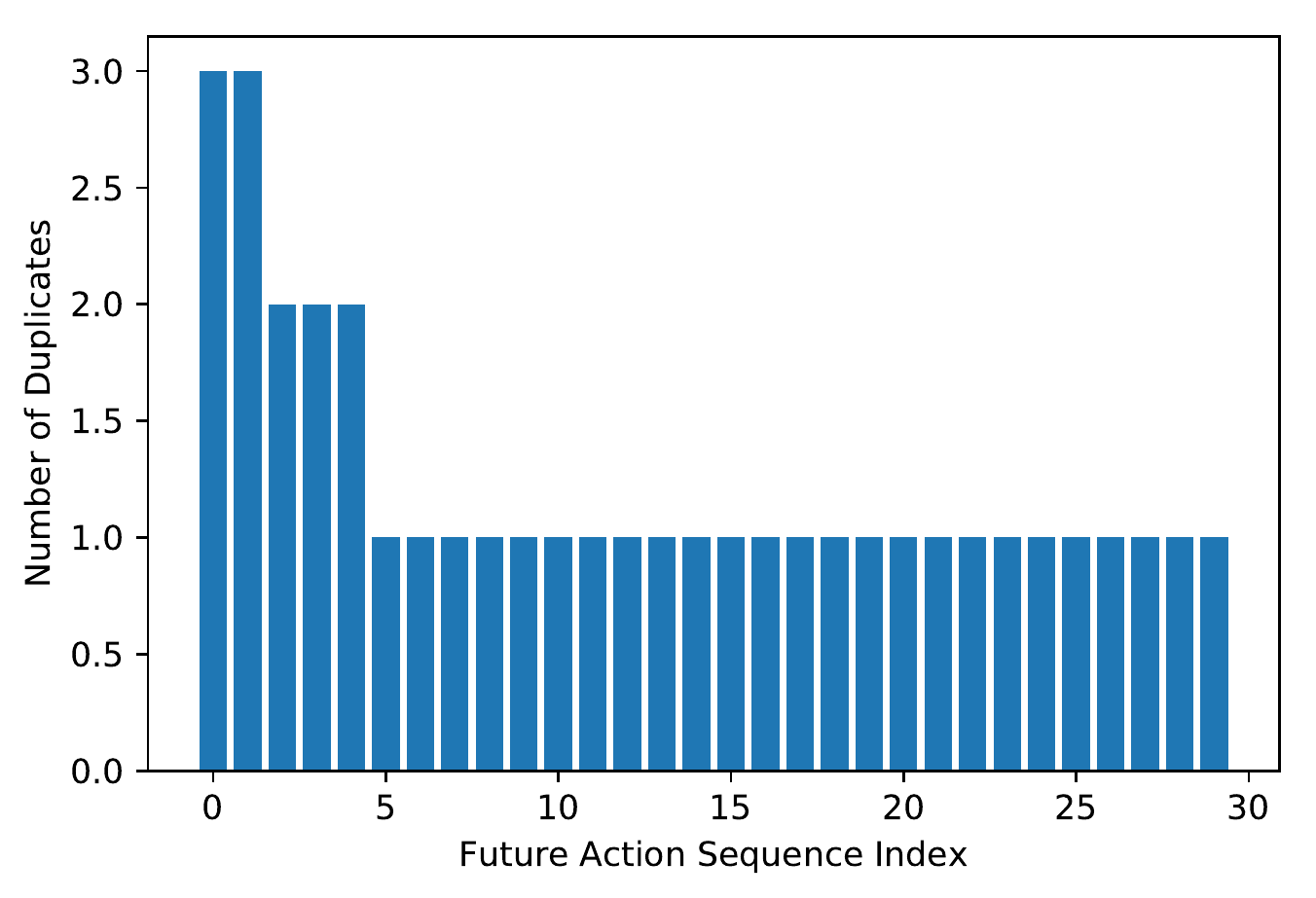}
    \caption{An example of the future action sequence frequency distribution of the \textit{intent} "\textbf{S/He wants to satisfy my hunger}". There are 30 distinct future action sequences matching this intent.}
    \label{fig:satisfy_his_hunger_distribution}
\end{figure}

An example of future action sequences of a selected intent is given in Figure \ref{fig:satisfy_his_hunger_distribution}. All of these conclusions pose a challenge not only for the generalization of multimodal matching, but also for compositional generalization. 

\subsection{Deep Generative Models Details}
\label{appendix:deep_generative}
We mainly adapt the multimodal deep planning model \uniVL to tackle our task. The training set of \uniVL consists of 2,402 videos, each of which contains a video clip of the initial state $\vs$, an observed action sequence $\va^i$, an intent $\vg$, and a future action sequence $\va^f$. Both models are trained to minimize prediction errors of $\va^f$. 

\paragraph{\uniVL} \uniVL~\cite{luo2020univl} is a SOTA unified pretrained vision-language model for multimodal understanding and generation. We consider \uniVL because \uniVL still performs the best on video captioning tasks, such as YouCook2 \cite{zhou2017automatic}. YouCook2 contains task-oriented and instructional third-person videos about indoor cooking. The captions of a video are provided for the whole video without explicit alignments at the frame or segment levels. In addition, \uniVL considers two sources of textual inputs: transcripts and captions. Hence, it is most close to our target task. Taking as input a future-directed intent and a video clip of the initial state, \uniVL is fine-tuned to forecast future action sequences.

More specifically, we utilize \uniVL to map a video clip to a sequence of action names. Most of the action names are multi-word expressions. During training, \uniVL takes as input both the visual features of a video clip $\vs$ and an observed action sequence $\va^i$, and optimizes the model with multiple pre-training objectives. The visual features are extracted by the \itd model~\cite{i3d} trained on \charades. During prediction, the model generates a future action sequence by only taking an initial visual state and high-level intent as input. To fine-tune \uniVL, we set the max.\ frame, mean frame and feature frame rate of the encoded features to be 629, 113 and 3. We fine-tune \uniVL on two NVIDIA V100 GPUs for 50 epochs and choose the best one based on the BLEU-3 metric.

\paragraph{Two Stage Planning Model.} The two stage planning baseline, \textbf{TwoStagePlan} for short, starts by converting the initial visual state $\vs$ into a textual description of the observed action sequence, followed by applying a Seq2Seq language model, \prophetNet \cite{qi2020prophetnet}, to predict future actions. 

At Stage 1, we adopt \uniVL on the video captioning task. Different from the single \uniVL baseline, we only train it with observed video clip inputs and let it generate the corresponding captions for observed action sequences. The other settings and training settings remain the same as for the single \uniVL baseline.

Given an observed action sequence recognized by \uniVL, we fine-tune \prophetNet by following~\newcite{jansen2020visuallyPlanning} in Stage 2. We prefer \prophetNet over GPT2~\cite{radford2019language} because it can learn to predict $n$ future tokens jointly, which is computationally efficient and mitigates overfitting on strong local correlations. For each video, we take as input the intent and the observed action sequence, separated by a special token \textit{SEP}, and train the model to minimize prediction errors of future action sequences. Fine-tuning the model from the \prophetNetEN pretrained checkpoint for 50 epochs on 2 Nvidia Tesla V100 GPUs, we choose the best model based on the validation loss.

\paragraph{\prophetNet} To study the impact of visual information, we consider a text-only baseline by employing \prophetNet. Herein, \prophetNet takes as input an intent and generates the future action sequences. The training is done with the same training procedure as Stage 2 of \UniVLProphetNet. This model serves for an ablation study, in contrast to \UniVLProphetNet, which uses additionally recognized action sequences as input. 

\subsection{Neurosymbolic Planning Model}
\label{appendix:neurosymbolic}
\label{5.1}
\begin{figure}[h]
    \centering
    \includegraphics[width=\linewidth,height=0.5\linewidth]{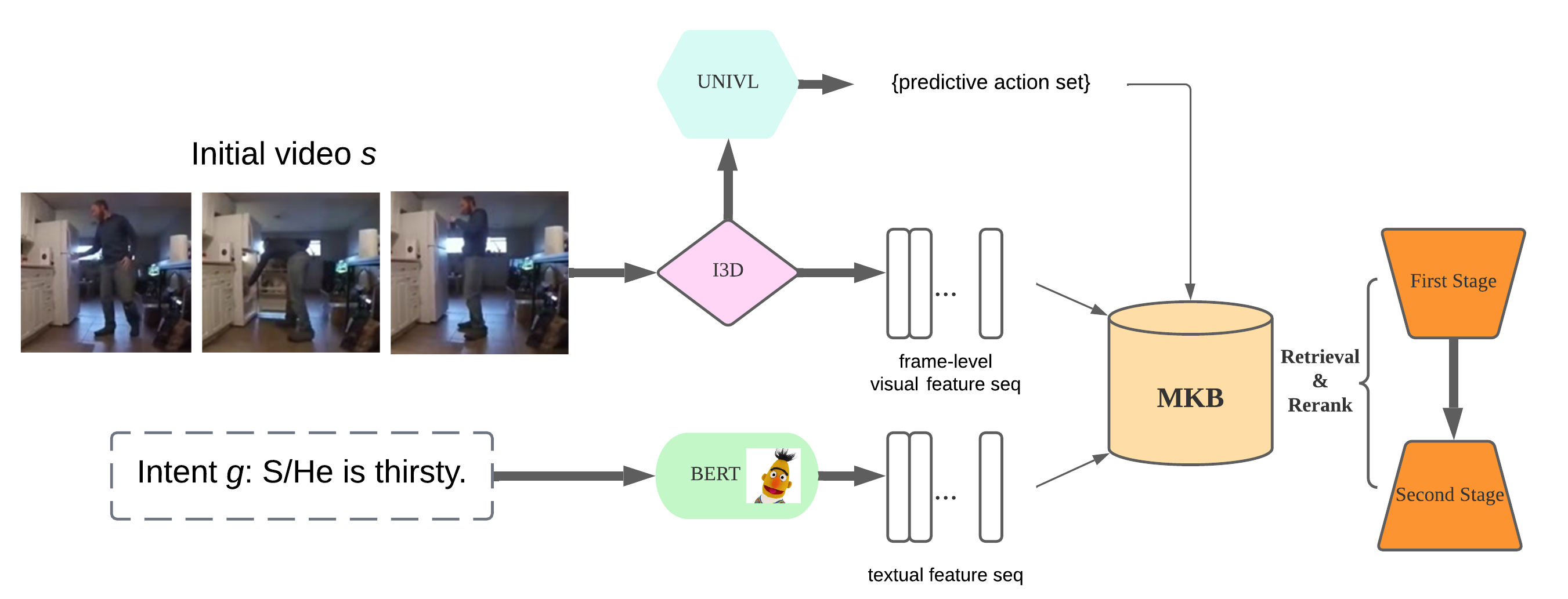}
    \caption{The neurosymbolic planning model is a multimodal retrieval \& re-rank pipeline.}
    \label{fig:retrieval}
\end{figure}
Instead of using the data in the training set to directly optimize model parameters, the neurosymbolic planning model (\textbf{NSPlan}) builds an MKB from the training data. Given a question in the test set, the model retrieves relevant knowledge based on the initial visual state and the intent, and then applies the retrieved knowledge to infer the most plausible answers from all available choices. 




\subsubsection{Retrieval from Multimodal Knowledge Base}
\label{sec:retrieval methods}
The neurosymbolic planning model retrieves relevant action sequences from the MKB in two stages. The first stage aims to computationally efficiently obtain all relevant action sequences. At the second stage, it re-ranks the initial retrieval results by using both visual and symbolic knowledge.  

\paragraph{First Stage.} Given the initial state of a video, we apply the pretrained \uniVL model used in the two-stage planning model to predict a sequence of observed actions. Then this action sequence in text form is sent as query to retrieve top-$50$ relevant action sequences from the MKB. For each retrieved result, the ranking score is the weighted sum of the \bmTF~\cite{robertson1994bm25} score between two action sequences and the cosine similarity between the intents. At this stage, only textual information is taken into account, and the temporal order of actions in a sequence is not considered because \bmTF considers each action sequence as a bag of words. 
\paragraph{Second Stage.} We re-rank the results from the first stage by taking temporal order and the visual features of action sequences into account. Each action sequence is represented as a sequence of frame-level visual feature vectors, which are extracted by the same visual encoder \itd. We apply the Ordered Temporal Alignment Module (OTAM) \cite{cao2020few} to compare two visual feature sequences. OTAM computes a distance between a pair of sequences by integrating video segment distances only along the ordered temporal alignment path. We turn a distance into an alignment score by $s_\mathrm{align} = 1  /  (1 + d_\mathrm{otam})$, where $d_\mathrm{otam}$ denotes the OTAM distance.

Many retrieved action sequences do not contain future actions. In order to rank the sequences with potential future actions higher, we add a rule to encourage long sequences containing unseen actions. The rule score $s_\mathrm{rule} = s_\mathrm{last} + s_\mathrm{len}$ is the sum of two binary indicator functions $s_\mathrm{last}$ and $s_\mathrm{len}$, where $s_\mathrm{last} = 1$ if and only if the last action of the retrieved result is not contained in the query set, and $s_\mathrm{len} = 1$ if and only if the length of the retrieved result is greater than that of the query. The final ranking score $s_f(r)$ of a result $r$ is the weighted sum of the initial ranking score, the alignment score $s_\mathrm{align}$ and the rule-driven score $s_\mathrm{rule}$. 
To reduce noise, we keep only the top-$10$ results for probabilistic reasoning.
\label{appendix:future_action_sequence}
We provide a completed version of the comparison among all baselines on future sequence evaluation in Table \ref{appendix:table: Comparison of top-K future sequences of two baselines}.
\begin{table*}[ht]
\begin{center}
 \adjustbox{max width=\textwidth}{\begin{tabular}{|c|c|c|c|c|c|c|c|c|c|}
\hline
 \multirow{2}{*}{setting} & \multicolumn{7}{c|}{Quality} & \multicolumn{2}{c|}{Diversity} \\ \cline{2-10} 
\multicolumn{1}{|c|}{} & precision & \multicolumn{1}{c|}{recall} & \multicolumn{1}{c|}{seq-item-acc} & seq-hits@5 & seq-hits@10 & BLEU-1 & BLEU-2 & \multicolumn{1}{c|}{Dist1} & Dist2 \\ 
\hline
\UniVLProphetNet &21.59 &15.59&9.26&10.00&\textbf{16.86}&12.50&3.58&32.55&58.54 \\\hline
\RetrievalScoring
&20.73&21.66&8.74&5.69&7.65&19.25&\textbf{6.80}&38.46&66.34\\\hline
\prophetNet
&21.35 &19.75 &8.12 &9.61&10.59& 18.66 & 5.52 & \textbf{51.96} &\textbf{81.93}\\\hline
\uniVL
&\textbf{23.67} &\textbf{22.02} &\textbf{9.71}  &\textbf{12.75}&16.08& \textbf{20.52} & 6.52 &47.10 &77.42\\\hline
\end{tabular}}
\caption{Comparison of top-$k$ future sequences of all systems.}
\label{appendix:table: Comparison of top-K future sequences of two baselines}
\end{center}
\end{table*}

\subsection{Full Action Sequence Comparison}
\begin{table*}[!ht]
\begin{center}

 \adjustbox{max width=\textwidth}{\begin{tabular}{|c|c|c|c|c|c|c|c|c|c|}
\hline
 \multirow{2}{*}{setting} & \multicolumn{7}{c|}{Quality} & \multicolumn{2}{c|}{Diversity} \\ \cline{2-10} 
\multicolumn{1}{|c|}{} & precision & \multicolumn{1}{c|}{recall} & \multicolumn{1}{c|}{seq-item-acc} & seq-hits@5 & seq-hits@10 & BLEU-1 & BLEU-2 & \multicolumn{1}{c|}{Dist1} & Dist2 \\ 
\hline
\UniVLProphetNet &38.71&30.73&11.06&0.59&1.37&29.60&13.71&15.45&35.37 \\\hline
\RetrievalScoring
&\textbf{41.63}&\textbf{35.81}&\textbf{11.46}&\textbf{5.69}&\textbf{8.43}&\textbf{34.14}&\textbf{15.90}&\textbf{28.03}&\textbf{62.08}\\\hline
\end{tabular}}
\caption{Comparison of top-$10$ full action sequences of all systems.}
\label{appendix:table: Comparison of top-K action sequences of two baselines}
\end{center}
\end{table*}
\subsection{Metrics}
\label{appendix:metrics}
\begin{itemize}
\item Seq-item-acc: Sequence item classification accuracy evaluates the exact action matching of the predicted action sequence with the ground truth, counting how many times the action in the predicted sequence matches the ground truth at the exact position. For top-10 sequences, we calculate the mean accuracy of all sequences.
\item Precision and recall: The precision and recall do not consider the order of ground truth. They both treat the actions inside the sequence as a unified set. The precision of top-10 sequences is computed by averaging the precision of each sequence, which measures the number of true actions over the number of total actions in the sequence. Here, we define the true action as the action that occurred in the ground truth. Similarly, the recall of top-10 sequences is also computed by averaging all sequences' recall, which is a measure of the true actions over the number of ground truth actions.
\item Seq-hit@$k$ Rate: The seq-hits scores measure the exact sequence matches, calculated as the number of examples whose top-$k$ sequences include the ground truth sequence, and we report the seq-hits@5 and seq-hits@10 accordingly. As for the retrieval-based baseline, we only consider the in-domain situation where the ground truth sequences have also appeared in the knowledge base.
\item BLEU: We use the standard BLEU-1 and BLEU-2 scores that are widely used in the Machine Translation Field and adapt them to our setting by computing the action-level match.
\item Dist: We report Dist1 (Distinct-1) and Dist2 (Distinct-2) following the standard definition \cite{distinct}, to measure the diversity of action sequences, based on the number of distinct $N$-gram of top-$10$ sequences.
\end{itemize}

\subsection{Full Table of Future Sequence Evaluation}

In Table \ref{appendix:table: Comparison of top-K action sequences of two baselines}, we compare \UniVLProphetNet with \RetrievalScoring, where both models are designed to output the full action sequence including the observed actions. It turns out that \RetrievalScoring performs consistently across all  metrics, indicating that \RetrievalScoring has a stronger ability to identify the most similar full action sequences in the MKB and training set. 

\subsection{Wilcoxon's signed-rank test}
\label{appendix:wilcoxon_test}
Wilcoxon's signed-rank test is a statistical hypothesis test used either to test the ranking of a set of samples or to compare the rankings of two populations using a set of matched samples. The calculated Wilcoxon signed-rank test $t$ value is 55.5 with a $p$ value of 0.7979, which shows that there is no significant difference between the two sets of human evaluation samples.

\subsection{Other Research Questions}
\label{appendix:other_rqs}
\paragraph{\textbf{How useful are symbolic, neural, or neurosymbolic knowledge?}} 

The goal of reasoning is to perform the probabilistic inference $\argmax_{c_i} p(c_{i} \mid \vs, \vg)$ over all possible answers. One of the key differences of \RetrievalScoring from the two generative models is that it introduces the time series similarity TWED to compare the future action sequences with each answer.

To understand the effects of TWED in the probabilistic reasoning module of our retrieval-based baseline, we compare it with three other similarity measures: (a) cosine similarity between the mean vectors of two sequences, (b) cosine similarity between the max-pooling results of two sequences, (c) the time series distance function Dynamic Time Warping (DTW) \cite{muller2007dynamic}. All of them are evaluated based on the same best performing top-10 retrieval results.  

We also evaluate different types of symbolic, neural, and neurosymbolic features used for computing action-level distance inside those measures: (a) action class ID, (b) the visual prototype features in the MKB, (c) the textual prototype features in the MKB, (d) concatenation of the visual prototype features and textual prototype features. 
\begin{table}[]
\centering


\begin{adjustbox}{width=\columnwidth}
\begin{tabular}{|c|c|c|c|c|}
\hline
\multicolumn{1}{|c|}{\multirow{2}{*}{Method}} & Action ID & Visual-proto & Text-proto & Visual + text proto \\ \cline{2-5} 
\multicolumn{1}{|c|}{} & Accuracy(\%) & \multicolumn{1}{c|}{Accuracy(\%)} & \multicolumn{1}{c|}{Accuracy(\%)} & \multicolumn{1}{c|}{Accuracy(\%)} \\ \hline
Mean & 53.33 & 60.19 & 46.82 & 48.42 \\
Max-Pooling & 43.92 &43.52  & 41.76 & 42.35 \\
DTW & 48.82 & 61.37 & 50.98 & 52.15 \\
TWED & 44.50 & \textbf{64.11} &  48.23& 50.19 \\ \hline
\end{tabular}
\end{adjustbox}
\caption{Reasoner-scoring performance with varying combinations of similarity measures and action-level features.}
\label{table:sim_features_reasoning}
\end{table}

It is clear from Table \ref{table:sim_features_reasoning} that TWED using the visual prototype features performs the best. The performance of the two time series metrics are comparable. Combining visual prototype features and textual prototypes features actually harms the performance. This is in contrast to the retrieval evaluation, which finds the symbolic representations most useful. This highlights the flexibility of this hybrid neurosymbolic system, which naturally supports choosing the most appropriate types of information for its respective modules.

We also experiment with the symbolic knowledge described by the STRIPS language. More specifically, we implement a symbolic planner based on STRIPS, which is able to check to what degree each answer is compatible with the preconditions and effects defined for each action class. Such symbolic knowledge can boost the overall accuracy of \RetrievalScoring to 82\% if we substitute the ground truth actions for the action sequences recognized by \uniVL. However, if we only use the predictions of \uniVL, which has both precision and recall around 32\%, the overall accuracy drops by almost 10\%.

\end{document}